\documentclass[11pt,twocolumn]{article}
\usepackage[utf8]{inputenc}
\usepackage{amsmath,amssymb}
\usepackage{graphicx}
\usepackage{url}
\usepackage{hyperref}
\usepackage{natbib}
\usepackage{booktabs}
\usepackage{caption}
\usepackage{subcaption}

\usepackage[margin=1in]{geometry}

\title{Advancing Retrieval-Augmented Generation for Structured Enterprise and Internal Data}
\author{Chandana Cheerla \\ IIT Roorkee \\ chandana\_c@mfs.iitr.ac.in}

\begin{document}

\maketitle

\begin{abstract}
Organizations increasingly rely on proprietary enterprise data—including HR records, structured reports, and tabular documents—for critical decision-making. While Large Language Models (LLMs) exhibit strong generative capabilities, they remain constrained by static pretraining, limited context windows, and challenges in processing heterogeneous data formats. Although conventional Retrieval-Augmented Generation (RAG) frameworks solve some of these constraints, they often fall short in handling structured and semi-structured data effectively.  
This work introduces an advanced RAG framework that combines hybrid retrieval strategies leveraging dense embeddings (all-mpnet-base-v2) and BM25, enhanced through metadata-aware filtering using SpaCy NER and cross-encoder reranking for improved relevance. The framework employs semantic chunking to preserve textual coherence and explicitly retains the structure of tabular data, ensuring the integrity of row-column relationships. Quantized indexing techniques are integrated to optimize efficiency, while a human-in-the-loop feedback mechanism and conversation memory enhance the system’s adaptability over time.  
Evaluations on proprietary enterprise datasets demonstrate significant improvements over baseline RAG approaches, with a 15\% increase in Precision@5 (90\% vs. 75\%), a 13\% gain in Recall@5 (87\% vs. 74\%), and a 16\% improvement in Mean Reciprocal Rank (0.85 vs. 0.69). Qualitative assessments further validate the framework’s effectiveness, showing higher scores in Faithfulness (4.6 vs. 3.0), Completeness (4.2 vs. 2.5), and Relevance (4.5 vs. 3.2) on a 5-point Likert scale. These findings underscore the framework’s capability to deliver accurate, comprehensive, and contextually relevant responses for enterprise knowledge tasks. Future work will focus on extending the framework to support multimodal data and integrating agent-based retrieval systems.
The source code will be progressively released at: \href{https://github.com/CheerlaChandana/Enterprise-Chatbot}{GitHub Repository}.

\end{abstract}
% \noindent

\section{Introduction}
Large Language Models (LLMs) such as GPT-4 \citep{openai2023gpt4}, PaLM \citep{chowdhery2022palm}, and LLaMA \citep{touvron2023llama} have significantly advanced capabilities in natural language understanding and generation, excelling in tasks like question answering, summarization, and knowledge retrieval. Despite these achievements, LLMs remain inherently constrained by static pretraining on fixed corpora and limited context windows \citep{bommasani2021opportunities,press2022train}, which bounds their adaptability to dynamic or proprietary enterprise data. In domains like corporate governance, human resources, and finance, critical information is often encapsulated in structured records, policy documents, and tabular formats that LLMs cannot naturally ingest or reason over post-deployment.  
Retrieval-Augmented Generation (RAG) frameworks have emerged to address this gap by integrating retrieval mechanisms with LLMs, enabling models to fetch external knowledge at inference time \citep{lewis2020retrieval,izacard2020leveraging}. This paradigm enhances response relevance and factuality by grounding generation in up-to-date, domain-specific data. However, conventional or baseline RAG methods, optimized primarily for unstructured textual data, face notable challenges when applied to enterprise datasets comprising structured, semi-structured, and tabular information \citep{guu2020realm}.  
Key limitations include:  
\begin{itemize}
    \item Fragmented Contextual Representation: Baseline chunking strategies, typically using fixed token-length splits, often fracture meaningful contexts, especially in complex documents like policies or manuals \citep{phang2021clustering}.
    \item Inadequate Handling of Tabular Data: Flattening tables into linear text formats destroys the intrinsic row-column relationships necessary for precise data retrieval within tables \citep{zhang2020turl}.
    \item Limited Retrieval Completeness: Exclusive reliance on dense embeddings or sparse keyword methods like BM25 \citep{robertson2009probabilistic} restricts the model’s ability to balance semantic understanding with exact matching.
    \item Absence of Relevance Reordering: Without reranking mechanisms, initial retrieval results may not adequately prioritize the most relevant information \citep{karpukhin2020dense}.
    \item Static Query Interpretation: These systems lack the capability to refine or disambiguate queries dynamically, resulting in suboptimal retrieval for vague or incomplete user prompts.
\end{itemize}

\subsection{Contributions}
To address these failings, we introduce an enterprise-focused RAG framework that systematically enhances the retrieval and generation process across diverse data modalities. The contributions of our system are as follows:  
\begin{enumerate}
    \item Hybrid Retrieval Mechanism: We combine dense semantic retrieval using all-mpnet-base-v2 \citep{reimers2019sentence} with BM25-based sparse retrieval, weighted at a 0.6 to 0.4 ratio, to synergize semantic depth with lexical accuracy.
    \item Structure-Aware Chunking: For text, we adopt a recursive character-based chunking strategy with empirically tuned chunk sizes (700 tokens) that balance coherence and granularity. For tabular data, we employ Camelot \citep{camelot} and Azure Document Intelligence to extract and store tables as structured JSON with metadata capturing file name, row identifiers, and column headers, enabling row-level indexing via FAISS for fine-grained retrieval.
    \item Metadata-Driven Filtering: Named Entity Recognition (NER) via spaCy \citep{spacy} enriches document metadata, facilitating entity-aligned filtering that sharpens retrieval relevance.
    \item Contextual Re-Ranking: We integrate a cross-encoder reranking layer using MS MARCO fine-tuned models \citep{nogueira2019passage}, which reorders retrieved candidates based on contextual alignment with the query.
    \item Interactive Query Refinement: Leveraging LLaMA and Mistral models on ChatGroq, our system supports query expansion and rephrasing, guided by user feedback and conversational memory to iteratively enhance retrieval and response quality.
\end{enumerate}
We evaluated our framework on a corpus of publicly available enterprise policy documents, including the HR Policy Dataset and analogous corporate materials. For tabular data, we experimented with both full-table chunking and row-level indexing, finding the latter significantly enhances precision in row-specific queries while the former suffices for small tables.

\subsection{Results}
Our approach delivers substantial improvements over baseline RAG systems:  
\begin{itemize}
    \item Precision@5: 90\% vs. 75\%
    \item Recall@5: 87\% vs. 74\%
    \item Mean Reciprocal Rank (MRR): 0.85 vs. 0.69
\end{itemize}
These results underscore the robustness and applicability of our framework in real-world enterprise contexts, delivering superior retrieval accuracy, comprehensive responses, and higher contextual relevance.

We envision extending this framework towards agentic RAG systems \citep{gao2022precise,shinn2023reflexion}, where intelligent agents autonomously select retrieval strategies, adaptively reformulate queries, and integrate multimodal data sources—including images, scanned documents, and audio—thereby broadening the landscape of enterprise knowledge augmentation.

\subsection{Paper Organization}
The rest of the paper is organized as follows:  
\begin{itemize}
    \item Section 2: Elaborates on the related work.
    \item Section 3: Presents the methodology.
    \item Section 4: Discusses experiments and results.
    \item Section 5: Concludes with directions for future work, including potential extensions towards agentic retrieval systems.
\end{itemize}

\section{Related Work}
Retrieval-Augmented Generation (RAG) has emerged as a pivotal paradigm to bridge the limitations of static LLMs by dynamically fetching relevant information at inference time. Lewis et al. \citep{lewis2020retrieval} introduced the foundational RAG architecture by coupling dense retrievers with generative models, improving factual grounding in open-domain question answering. Building on this, Izacard and Grave \citep{izacard2020leveraging} proposed Fusion-in-Decoder (FiD), integrating multiple retrieved contexts directly into the decoding process, significantly enhancing response accuracy.  
Dense retrieval techniques such as Dense Passage Retrieval (DPR) \citep{karpukhin2020dense} rely on bi-encoder models to map queries and documents into the same vector space, enabling efficient retrieval via vector similarity. However, DPR and similar dense retrievers can struggle with lexical precision, especially when queries are domain-specific or contain rare terminology. Sparse retrieval methods like BM25 \citep{robertson2009probabilistic} compensate for this by focusing on exact term matching, which, while precise, often lacks semantic understanding.  
To overcome the trade-offs between dense and sparse retrieval, hybrid strategies have been explored. Mialon et al. \citep{mialon2023augmented} surveyed augmented language models and emphasized that combining dense and sparse signals leads to better retrieval performance across varied data types. Yet, most hybrid retrieval approaches have predominantly targeted unstructured textual data, with limited exploration into structured or semi-structured formats such as tables and structured records.  
Handling structured data directly has been tackled through models like TURL (Table Understanding through Representation Learning) by Zhang et al. \citep{zhang2020turl}, which encodes tabular semantics for downstream tasks, and TAPAS \citep{herzig2020tapas}, which leverages transformers to answer questions over tables. However, these models are optimized for direct table question answering rather than integrating into broader retrieval-generation pipelines that handle mixed data formats.  
Reranking retrieved candidates post-retrieval further enhances relevance. Nogueira and Cho \citep{nogueira2019passage} introduced BERT-based passage reranking, showing that cross-encoders can significantly refine retrieval results by evaluating query-document pairs holistically. Similarly, ColBERT \citep{khattab2020colbert} employs a late interaction mechanism, balancing between efficiency and precision, though its scalability remains an area of active research.

\subsection{Advantages of Prior Approaches}
\begin{itemize}
    \item Semantic and Lexical Balance: Hybrid retrieval effectively combines semantic depth with keyword accuracy.
    \item Improved Relevance: Reranking mechanisms like cross-encoders enhance the contextual fit of retrieved results.
    \item Handling Structured Data: Specialized models like TAPAS \citep{herzig2020tapas} and TURL \citep{zhang2020turl} address the nuances of tabular understanding.
\end{itemize}

\subsection{Gap Addressed by Our Work}
While these advancements have laid a robust foundation, there remains a lack of unified frameworks that seamlessly integrate hybrid retrieval, structured data handling, metadata filtering, reranking, and interactive refinement specifically tailored for enterprise environments. Our work addresses this comprehensive need, extending RAG capabilities to structured, semi-structured, and unstructured enterprise data, setting a precedent for future multimodal and agent-driven retrieval systems.

\section{Methodology}
 % Or \newpage to ensure text starts after the figure

Our proposed advanced RAG framework is designed to effectively retrieve and generate responses from heterogeneous enterprise data, including text, structured documents, and tabular records. The architecture addresses limitations in naive RAG pipelines through a combination of optimized document preprocessing, hybrid retrieval strategies, advanced ranking mechanisms, and feedback-driven refinement. This section details each stage of our system.

\subsection{Document Preprocessing and Chunking}
\subsubsection{Text Extraction and Semantic Chunking}
All textual documents, primarily sourced from publicly available HR policies (e.g., NASSCOM datasets), are extracted using pdfplumber. The extracted text is segmented using a Recursive Character Text Splitter with a chunk size of 2000 characters and a 500-character overlap, ensuring semantic coherence while maintaining the token constraints of LLMs.

\subsubsection{Table Extraction and Representation}
To accurately handle tabular data, we employ Camelot \citep{camelot} for table detection and extraction. Tables are serialized into JSON format, capturing metadata such as:  
\begin{itemize}
    \item File Name
    \item Row and Column Identifiers
    \item Cell Values
    \item Split into individual rows, each indexed separately to facilitate row-level retrieval.
\end{itemize}
When Camelot is insufficient (e.g., for complex formatting), we fallback to pdfplumber for extracting table-like structures. This ensures robust extraction across diverse document types.

\subsection{Metadata Enrichment}
We apply spaCy’s Named Entity Recognition (NER) \citep{spacy} to annotate each chunk with entities such as locations, dates, and organizational names. Additional metadata like document type, department, and confidentiality level is simulated for experimentation but can be replaced with real enterprise metadata.

\subsection{Hybrid Retrieval Strategy}
Our retrieval pipeline combines dense retrieval, sparse retrieval, and reranking, enhancing both semantic understanding and keyword precision.

\subsubsection{Dense Retrieval}
Chunks are embedded using all-mpnet-base-v2 embeddings \citep{reimers2019sentence}, a state-of-the-art sentence embedding model. These embeddings are stored in a FAISS HNSW index (with M=32, efConstruction=200, efSearch=50), enabling efficient approximate nearest neighbor searches.

\subsubsection{Sparse Retrieval}
We employ BM25 \citep{robertson2009probabilistic}, a classic keyword-based retrieval algorithm, to complement dense retrieval, particularly beneficial for exact term matching.

\subsubsection{Retrieval Fusion}
Dense and sparse retrieval scores are combined using a weighted sum:  
\begin{equation}
\text{Score}_{\text{combined}} = 0.6 \times \text{Score}_{\text{dense}} + 0.4 \times \text{Score}_{\text{sparse}}
\end{equation}
This weighting was determined empirically to balance semantic and lexical relevance.

\subsection{Contextual Reranking}
The top candidate chunks are reranked using a Cross-Encoder reranker based on the ms-marco-MiniLM-L-12-v2 model \citep{nogueira2019passage}. This cross-encoder evaluates the query-chunk pairs to reassign relevance scores, ensuring that the most contextually aligned documents are prioritized.

\subsection{Query Formulation and Refinement}
To enhance initial queries, we incorporate:  
\begin{itemize}
    \item Query Rewriting: Rephrasing vague or incomplete queries using LLaMA or Mistral models on ChatGroq.
    \item Query Expansion: Generating alternative query formulations to cover broader aspects of the topic.
\end{itemize}
If a user flags an answer as unsatisfactory, the query is automatically expanded and retried, leveraging the LLM to guide reformulation.

\subsection{Answer Generation with LLMs}
Final reranked chunks are passed to LLMs for answer synthesis. We utilize:  
\begin{itemize}
    \item Mistral-7B and LLaMA models on ChatGroq for their balance between accuracy and inference speed.
    \item A Grounded Prompt Template that instructs the model to:
    \begin{itemize}
        \item Answer strictly based on retrieved sources.
        \item Use bullet points for clarity.
        \item Provide citations to source documents.
        \item Include summaries if the response exceeds three sentences.
    \end{itemize}
\end{itemize}

\subsection{Feedback Loop and Conversational Memory}
A ConversationBufferMemory retains up to 10 recent interactions, maintaining session continuity. Additionally, user feedback (thumbs up/down) is logged. Negative feedback triggers automated query reformulation and re-retrieval, enhancing system adaptivity over time.

\subsection{Index Optimization}
We maintain two FAISS indices:  
\begin{itemize}
    \item High-Precision Index: Uses all-mpnet-base-v2 embeddings \citep{reimers2019sentence} for general queries.
    \item Lightweight Index: Uses paraphrase-MiniLM-L3-v2 embeddings for resource-constrained environments.
\end{itemize}
This dual-index system offers a trade-off between computational efficiency and retrieval accuracy.

\subsection{Evaluation Dataset and Setup}
Our evaluations used a corpus of predominantly HR policies from publicly available sources and corporate reports. The dataset contains a rich mix of text and tables, simulating real enterprise data diversity.

\begin{figure}[htbp]
    \centering
    \includegraphics[width=\textwidth,height=16cm]{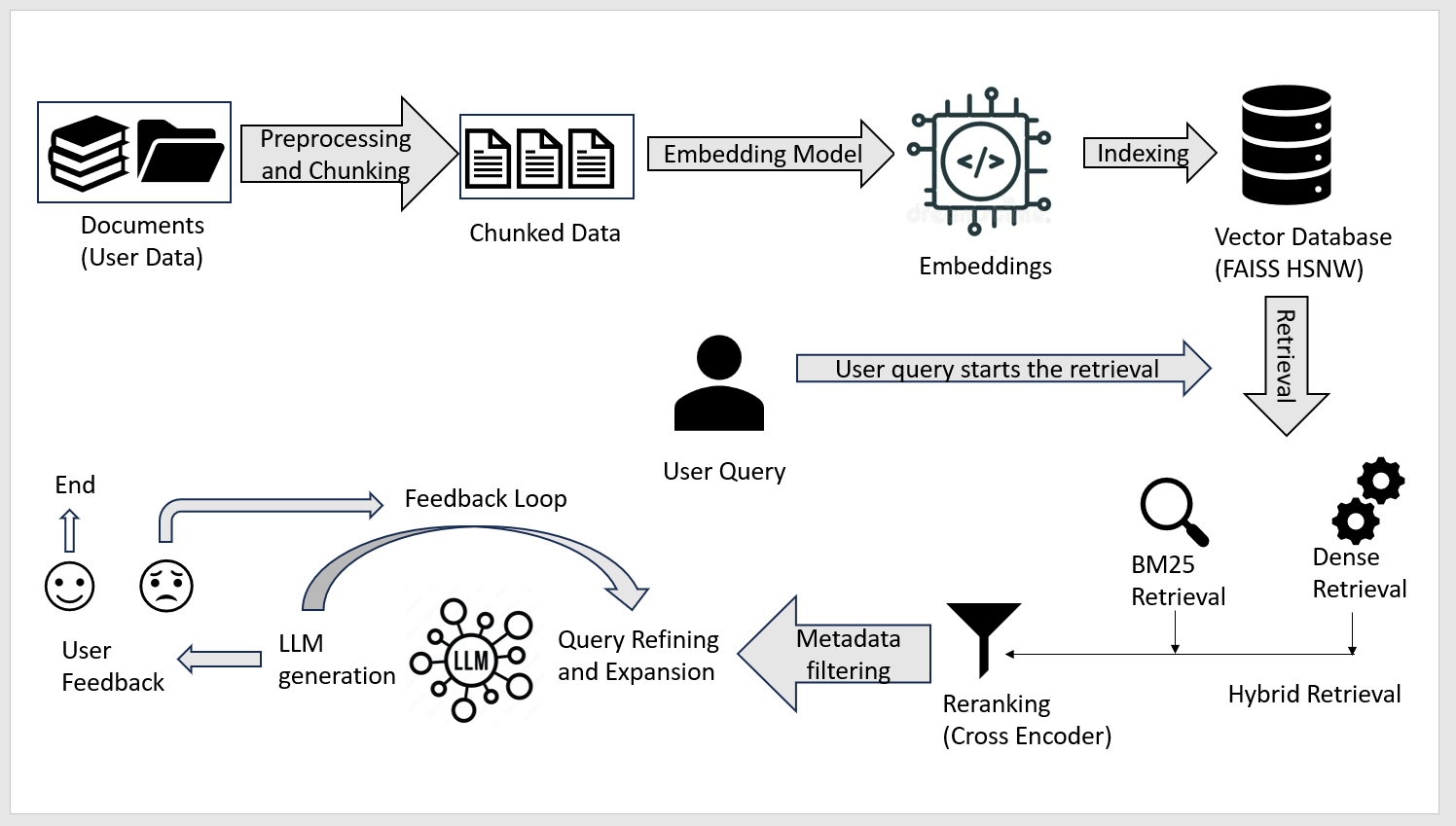}
    \caption{Architecture Diagram of the Proposed RAG Framework}
    \label{fig:rag-architecture}
\end{figure}

\clearpage

\subsection{Metrics}
We evaluated the system quantitatively and qualitatively:  
\begin{itemize}
    \item Precision@5, Recall@5, MRR: Standard retrieval metrics.
    \item Faithfulness, Completeness, Relevance: Assessed on a 5-point Likert scale by human evaluators.
\end{itemize}

\section{Experiments}
\subsection{Dataset}
We evaluated our framework on a dataset comprising publicly available HR policies (e.g., companies and institutions data ) and corporate reports. This dataset represents a diverse collection of unstructured text, structured data, and tabular content, simulating real-world enterprise knowledge repositories.Additionally, supplementary experiments utilized datasets from public repositories such as,
\url(https://www.data.gov.in/)
\url(https://archive.ics.uci.edu/ml/datasets/adult)
\href{https://github.com/purvikanani/HR_analytics}{HR Analytics GitHub Repository}.

\subsection{Experimental Procedure}
We adopted a progressive experimental methodology, beginning with a naive RAG baseline and incrementally integrating advanced retrieval and generation techniques. This stepwise approach allowed us to quantify the impact of each enhancement in the pipeline.

\subsection{Naive RAG Baseline}
The baseline system was configured as follows:  
\begin{itemize}
    \item Recursive Character-Level Chunking: Documents were segmented into chunks of 500, 700, and 1000 characters to balance context preservation and retrievability. A chunk size of approximately 700 characters yielded optimal performance, although variations across sizes were marginal.
    \item Dense Retrieval Only: Document embeddings were generated using all-mpnet-base-v2 \citep{reimers2019sentence}, indexed via FAISS for dense retrieval.
    \item Direct LLM Generation: Retrieved chunks were passed directly to the LLM without any reranking or filtering mechanisms.
    \item Tabular Data: Initially, tables were treated as plain text and chunked similarly to unstructured text, which led to suboptimal performance, especially for row-specific queries.
\end{itemize}

\subsection{Table Handling Strategies}
Recognizing the inadequacies in handling tabular data, we experimented with several strategies:  
\begin{enumerate}
    \item Storing Entire Tables as Chunks: Effective for small tables (less than 10 rows), but impractical for larger tables due to context window constraints and poor precision in row-specific retrievals.
    \item Azure Document Intelligence: Employed to parse tables into structured formats enriched with detailed metadata for rows, columns, and headers.
    \item Camelot Integration: Utilized Camelot \citep{camelot}, an open-source tool, to extract tables into JSON format, preserving structural relationships.
    \item Row-Level Indexing: Each table row was indexed separately within FAISS, enabling fine-grained retrieval for row-specific queries.
\end{enumerate}

\subsection{Full Advanced RAG Pipeline}
Upon refining the table handling mechanisms, we implemented the complete advanced RAG pipeline:  
\begin{itemize}
    \item Hybrid Retrieval: Combined dense retrieval (all-mpnet-base-v2 \citep{reimers2019sentence}) with sparse retrieval (BM25 \citep{robertson2009probabilistic}) using a weighted fusion: 0.6 (dense) and 0.4 (sparse).
    \item Semantic and Table-Aware Chunking: Enhanced chunking strategies ensured coherence for textual data and structural integrity for tables.
    \item Cross-Encoder Reranking: Applied ms-marco-MiniLM-L-12-v2 \citep{nogueira2019passage} as a cross-encoder reranker to score and reorder the top-k retrieved chunks.
    \item Query Refinement: Integrated automatic query rewriting and expansion using LLaMA and Mistral models on ChatGroq, especially in response to negative user feedback.
    \item Feedback Loop: A human-in-the-loop mechanism where user feedback triggered query reformulation and re-retrieval.
\end{itemize}

\subsection{Evaluation Metrics}
We assessed the system using both quantitative and qualitative metrics:  
\begin{itemize}
    \item Precision@5: The proportion of relevant documents within the top 5 retrieved results, indicating immediate utility.
    \item Recall@5: The proportion of all relevant documents captured within the top 5 results, reflecting coverage.
    \item Mean Reciprocal Rank (MRR): Measures the rank position of the first relevant document, with higher values indicating quicker retrieval.
\end{itemize}
Additionally, qualitative evaluations were conducted with human annotators who rated responses on:  
\begin{itemize}
    \item Faithfulness: The degree to which generated answers accurately reflect retrieved content without hallucination.
    \item Completeness: The extent to which the response comprehensively addresses the query.
    \item Relevance: The pertinence of the response to the user’s query.
\end{itemize}
And also we use LLMs as an another evaluator. Each qualitative metric was assessed on a 5-point Likert scale.

\subsection{Results}
\begin{figure}[htbp]
    \centering
    \includegraphics[width=0.5\textwidth,height=7cm]{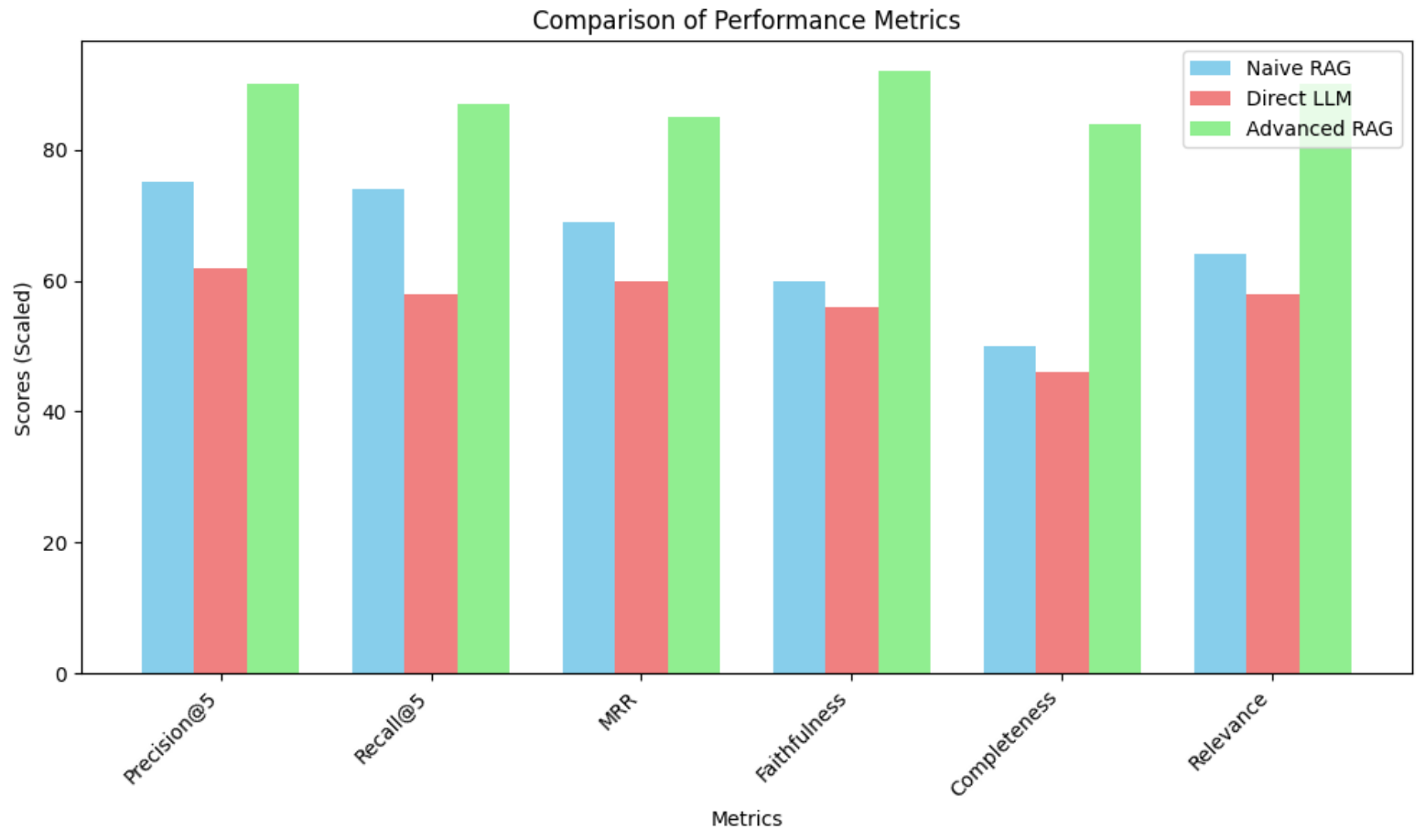}
    \caption{Comparision of the performance metrics}
    \label{fig:rag-architecture}
\end{figure}

\begin{table}[htbp]
    \centering
    \resizebox{0.5\textwidth}{!}{ % Adjust to 50% of the text width
    \begin{tabular}{lccc}
        \toprule
        Metric            & Direct LLM & Naive RAG & Advanced RAG \\
        \midrule
        Precision@5       & 62\%      & 75\%        & 90\%          \\
        Recall@5          & 58\%      & 74\%        & 87\%          \\
        MRR               & 0.60      & 0.69        & 0.85          \\
        Faithfulness      & 2.8       & 3.0         & 4.6           \\
        Completeness      & 2.3       & 2.5         & 4.2           \\
        Relevance         & 2.9       & 3.2         & 4.5           \\
        \bottomrule
    \end{tabular}
    }
    \caption{Comparison of performance metrics across Naive RAG, Direct LLM, and Advanced RAG.}
    \label{tab:results}
\end{table}

\subsection{Key Observations}
\begin{itemize}
    \item Chunking: A 700-character chunk size offered a balanced trade-off between context preservation and retrievability, though the impact of chunk size was not significant beyond certain thresholds.
    \item Table Handling: Implementing row-level indexing drastically improved retrieval precision for tabular queries compared to naive chunking.
    \item Component Contributions: Each enhancement—hybrid retrieval, cross-encoder reranking, and query refinement—provided incremental improvements, cumulatively resulting in substantial gains in both quantitative and qualitative metrics.
\end{itemize}

\subsection{Summary}
Our experimental results demonstrate that the proposed advanced RAG framework significantly outperforms both naive RAG and direct LLM prompting approaches. The combination of hybrid retrieval, semantic and structure-aware chunking, cross-encoder reranking, and dynamic query refinement enables effective handling of heterogeneous enterprise data, including complex tabular formats.  
Consistent improvements across Precision@5, Recall@5, and MRR, alongside higher human ratings for faithfulness, completeness, and relevance, affirm the robustness of our pipeline for real-world enterprise knowledge augmentation tasks. These findings validate the necessity of tailored retrieval strategies and structured data handling within enterprise RAG systems.

\section{Advantages of the Proposed Framework}
Our advanced RAG framework introduces several key improvements that make it well-suited for handling enterprise data retrieval and generation tasks.  
First, the framework is capable of effectively working with a wide variety of data formats commonly found in enterprises, including unstructured text, structured documents, and tabular data. This versatility makes it practical for real-world scenarios where information is dispersed across different formats.  
Second, the hybrid retrieval approach—combining dense embeddings with sparse keyword-based methods—strikes a balance between semantic understanding and exact keyword matching. This ensures that the system retrieves information that is both contextually relevant and factually precise. The additional layer of cross-encoder reranking further refines the results, prioritizing the most relevant content.  
Another strength of the framework is its approach to tabular data. By implementing table-aware chunking and indexing each row individually, the system achieves a level of granularity that allows it to answer row-specific queries more effectively than standard text chunking would allow.  
Additionally, the system includes dynamic query optimization through LLM-based rewriting and expansion, enabling it to refine ambiguous or incomplete queries. This makes the retrieval process more robust, especially when dealing with varied user inputs.  
For generation, the use of a grounded prompting strategy ensures that LLM responses remain anchored in the retrieved evidence, with citations and summaries provided where necessary. This not only enhances the credibility of the answers but also mitigates the risk of hallucinations.  
Lastly, the framework is designed with scalability in mind through a dual-index approach—offering a high-precision index for accuracy-demanding tasks and a lightweight alternative for resource-constrained environments. The system's modular design also makes it adaptable to domains beyond HR and corporate data, such as healthcare, legal, and financial applications.

\section{Limitations and Future Work}
While the framework offers several advantages, there are still some limitations that we aim to address in future work.  
One of the primary limitations is the reliance on static indexing. As it stands, any updates to the document corpus require full reindexing, which can be time-consuming and impractical in environments where data changes frequently. A more dynamic, incremental indexing mechanism is needed to make the system responsive to real-time data updates.  
Another area for improvement is in handling highly complex or nested tables. Although our current approach performs well on simple to moderately complex tables, it can struggle with preserving relationships in deeply structured tables with hierarchical headers or merged cells. This sometimes leads to partial loss of context during retrieval.  
The feedback mechanism, while useful, currently depends on explicit user input like thumbs up or down. In real-world settings, users may not always provide such feedback, limiting the system's capacity to learn and adapt automatically. Incorporating passive signals, such as how users interact with the retrieved information, could help the system improve over time without requiring direct input.  
We also see limitations in the query reformulation process. Presently, it relies on heuristic-based expansions using LLMs, which can occasionally misinterpret user intent or broaden the query too much, leading to less focused results. More intent-aware or interactive clarification methods could address this issue.  
Finally, while our retrieval fusion strategy is effective on moderately sized datasets, scaling it to very large corpora poses computational challenges, particularly in terms of retrieval latency and memory usage. More efficient fusion and reranking techniques will be essential for handling enterprise-scale deployments.

\subsection{Future Work}
Looking ahead, we plan to enhance the framework in several ways. One direction is to implement dynamic indexing capabilities, allowing the system to update its indices incrementally as new data arrives. We also intend to integrate advanced table understanding models like TAPAS \citep{herzig2020tapas} or TURL \citep{zhang2020turl}, which are specifically trained to preserve relational structures in complex tables.  
In terms of user feedback, we aim to explore methods for leveraging implicit feedback signals—such as click-through rates and time spent on documents—to guide system improvements without relying on explicit ratings.  
To improve query handling, we are considering the use of agent-based approaches, such as ReAct agents \citep{yao2022react}, that can reason about query intent and dynamically adapt retrieval strategies.  
Finally, we plan to extend the system to support multimodal data, including scanned documents, images, and charts, broadening its applicability across various enterprise contexts.

\bibliographystyle{plain}
\bibliography{references}
   % 'references' is the name of your .bib file (without .bib extension)

\end{document}